\documentclass[letterpaper, 10 pt, conference]{ieeeconf}
\IEEEoverridecommandlockouts
\usepackage[english]{babel}
\usepackage[pdftex]{graphicx}
\graphicspath{{./figures/}}
\DeclareGraphicsExtensions{.pdf,.jpg,.jpeg,.png}
\usepackage[cmex10]{amsmath}
\interdisplaylinepenalty=2500{}
\usepackage{booktabs,multirow}
\usepackage[caption=false,font=footnotesize]{subfig}
\def\enquote#1{\lq{#1}\rq}

\usepackage{tikz}
\usetikzlibrary{shapes}
\newcommand*\action[1]{ \scalebox{0.8}{\tikz[baseline=(char.base)]{\node[rectangle,draw,inner sep=2pt] (char) {#1};}} }
\newcommand*\faction[1]{ \scalebox{0.8}{\tikz[baseline=(char.base)]{\node[ellipse,draw,inner sep=0.5pt,text width=3.5mm,align=center] (char) {#1};}} }
\usepackage{url}
\newcommand{\fref}[1]{Fig.~\ref{#1}}
\newcommand{\tref}[1]{Table~\ref{#1}}
\newcommand{\sref}[1]{Section~\ref{#1}}

\newcommand\mat[1]{\boldsymbol{#1}}
\newcommand\vect[1]{\boldsymbol{#1}}
\newcommand\matop[2]{\boldsymbol{#1}\left({#2}\right)}
\usepackage[acronym,shortcuts]{glossaries}
\glsdisablehyper
\newacronym[longplural=Degrees of Freedom]{dof}{DoF}{Degree of Freedom}
\newacronym{ft}{F/T}{Force-Torque}
\newacronym{ros}{ROS}{Robot Operating System}
\newacronym{rrt}{RRT}{Rapidly-Exploring Random Tree}
\newacronym{wam}{WAM}{Whole Arm Manipulator}

\title{\LARGE \bf{}
A Framework for Fine Robotic Assembly
}

\author{Francisco Su\'{a}rez-Ruiz and Quang-Cuong Pham%
  \thanks{F. Su\'{a}rez-Ruiz and Q.-C. Pham are with the School of
    Mechanical and Aerospace Engineering, NTU, Singapore.}%
  \thanks{This work was supported by Tier $1$ grant RG$109/14$ awarded
    by the Ministry of Education of Singapore.}  }

\begin{document}
\maketitle
\thispagestyle{empty}
\pagestyle{empty}

\begin{abstract}
  Fine robotic assembly, in which the parts to be assembled are small and fragile and lie in an unstructured environment, is still out of reach of today's industrial robots. The main difficulties arise in the precise localization of the parts in an unstructured environment and the control of contact interactions. Our contribution in this paper is twofold. First, we propose a taxonomy of the manipulation primitives that are specifically involved in fine assembly. Such a taxonomy is crucial for designing a scalable robotic system (both hardware and software) given the complexity of real-world assembly tasks. Second, we present a hardware and software architecture where we have addressed, in an integrated way, a number of issues arising in fine assembly, such as workspace optimization, external wrench compensation, position-based force control, etc. Finally, we show the above taxonomy and architecture in action on a highly dexterous task -- bimanual pin insertion -- which is one of the key steps in our long term project, the autonomous assembly of an IKEA chair.
\end{abstract}

\section{Introduction}
\label{sec:introduction}

Robotics has largely contributed to increasing industrial productivity
and to helping factory workers on tedious, monotonous tasks, such as
pick and place, welding, or painting. There are however some major
challenges that still prevent the automation of many repetitive tasks
-- especially in \enquote{light} industries -- such as the assembly of
small parts in the electronics, shoes or food industries.

As opposed to \enquote{heavy} industries, where sophisticated assembly
lines provide a highly structured environment (for instance, on car
assembly lines, the position of the car frame is known to
sub-millimeter precision), \enquote{light} industries are associated
with unstructured environments, where the small parts to be assembled
are placed in diverse positions and orientations. While tremendous
progress has been made in 3D perception in recent years, current
3D-vision systems are still not precise enough for fine assembly.

Another related problem is that most robots currently used in the
industry are \emph{position-controlled}, that is, they can achieve
very precise control in position and velocity, at the expense of poor,
or no, control in force and torque. Yet, force or compliant control is
crucial while assembling fragile, soft, small parts. Assembly tasks
imply by essence contacts between the robot and the environment,
making the sensing and control of contact forces decisive. A number of
compliant robots have been developed in recent years, such as the KUKA
Lightweight Robot~\cite{Bischoff2010} or the Barrett
\ac{wam}~\cite{Rooks2006}, but, compared to existing industrial
robots, they are still one order of magnitude more expensive, less
robust and more difficult to maintain. We believe therefore that the
key to automatizing \enquote{light} industries lies in augmenting
existing industrial position-controlled manipulators with extra
functionalities, such as compliant control, through the addition of
affordable hardware components (e.g. end-effector force/torque sensor)
and smart planning, sensing and control software.


The goal of this paper is to present our framework dedicated to fine
assembly -- our long term project being to demonstrate the capability
of that framework by autonomously assembling an IKEA
chair. 

Previous works have attempted to complete similar tasks \cite{Lee1993,Knepper2013,Wahrburg2014}. Specifically, Knepper et al. \cite{Knepper2013} present a multi-robot system that assembles an IKEA table. They focus more on the task planning architecture than in the challenges of fine assembly. To cope with the force interactions they need a dedicated tool based on a compliant gripper for screwing the table legs. In order to use off-the-shelf components, we prefer software over mechanical compliance.

In this paper, we discuss two initial contributions. First, we
propose a taxonomy of the manipulation primitives involved in fine
assembly. Such a taxonomy serves as crucial guideline for designing a
scalable robotic manipulation system (both hardware and software),
given the complexity of real-world assembly tasks. In particular,
thinking in terms of primitives moves beyond the low-level
representations of the robot's movements (classically joint-space or
task-space) and enables generalizing robot capabilities in terms of
elemental actions that can be grouped together to complete any task.

As our taxonomy is tailored for industrial fine assembly, it differs
from existing manipulation
taxonomies~\cite{Cutkosky1990,Feix2009,Bullock2012,Owen-Hill2013} in
two key aspects: (i) we focus on parallel-jaw grippers (the most
common and robust gripper in the industry), which excludes some
complex primitives such as in-hand manipulation; (ii) in addition to
the interaction of the gripper with the gripped object, we also
consider \emph{multi-object interactions} (e.g. the gripped object
interacts with another object), which constitute the essence of
assembly.

Our second contribution is the development of a hardware and software
framework based on the above taxonomy and tailored for robotic
assembly. The hardware comprises an optical motion capture system and
two industrial position-controlled manipulators, each equipped with a
force/torque (F/T) sensor at the wrist and a parallel gripper. The two
manipulators are necessary since most assembly tasks require two hands
to complete (see~\cite{Smith2012} for a complete survey on bimanual
manipulation). Compared to integrated bimanual robots, such as the
Toyota Dual Arm Robot~\cite{Bloss2010}, the Yaskawa Motoman
SDA10D~\cite{Yamada1995}, or the ABB dual arm YuMI~\cite{Kock2011},
our two independent manipulators enable higher workload and larger
workspace, at a fraction of the cost.

On the software side, we address a number of issues arising in fine
assembly, such as workspace optimization, external wrench
compensation, position-based force control, etc. These issues have
often been discussed in the literature, but we address them here in
an integrated way and on a single software platform built on top of
the \ac{ros}~\cite{Quigley2009}. We intend to make this platform
available as open-source in the near future.

To illustrate the above developments, we consider a highly dexterous
task: bimanual pin insertion. This task requires most of the
capabilities just mentioned, such as bimanual motion planning, object
localization, control of contact interactions, etc. It also
constitutes one of the key steps in our long term project, the
autonomous assembly of an IKEA chair. Finally, it yields a fully
quantifiable way to measure the dexterous performance of a robotic
manipulation system and is therefore a good test for the
generalizability and simplicity of implementation.

The paper is structured as follows. In \sref{sec:primitives}, the
proposed manipulation taxonomy is presented. The manipulation
primitives have been selected for assembly tasks, but can also be used
for the description of any type of robotic tasks. \sref{sec:setup}
gives details regarding our hardware and software framework and
describes the requirements that we have identified as essential to
perform all the motion primitives. \sref{sec:task} depicts the
bimanual pin insertion task, showing how it can be broken down into
different subtasks, which in turn can be divided into manipulation
primitives. Finally, \sref{sec:conclusions} draws some conclusions and
sketches some directions for future research.

\section{Manipulation Primitives for Assembly}
\label{sec:primitives}
\begin{figure*}[t]
  \centering
  \includegraphics[width=0.98\textwidth ]{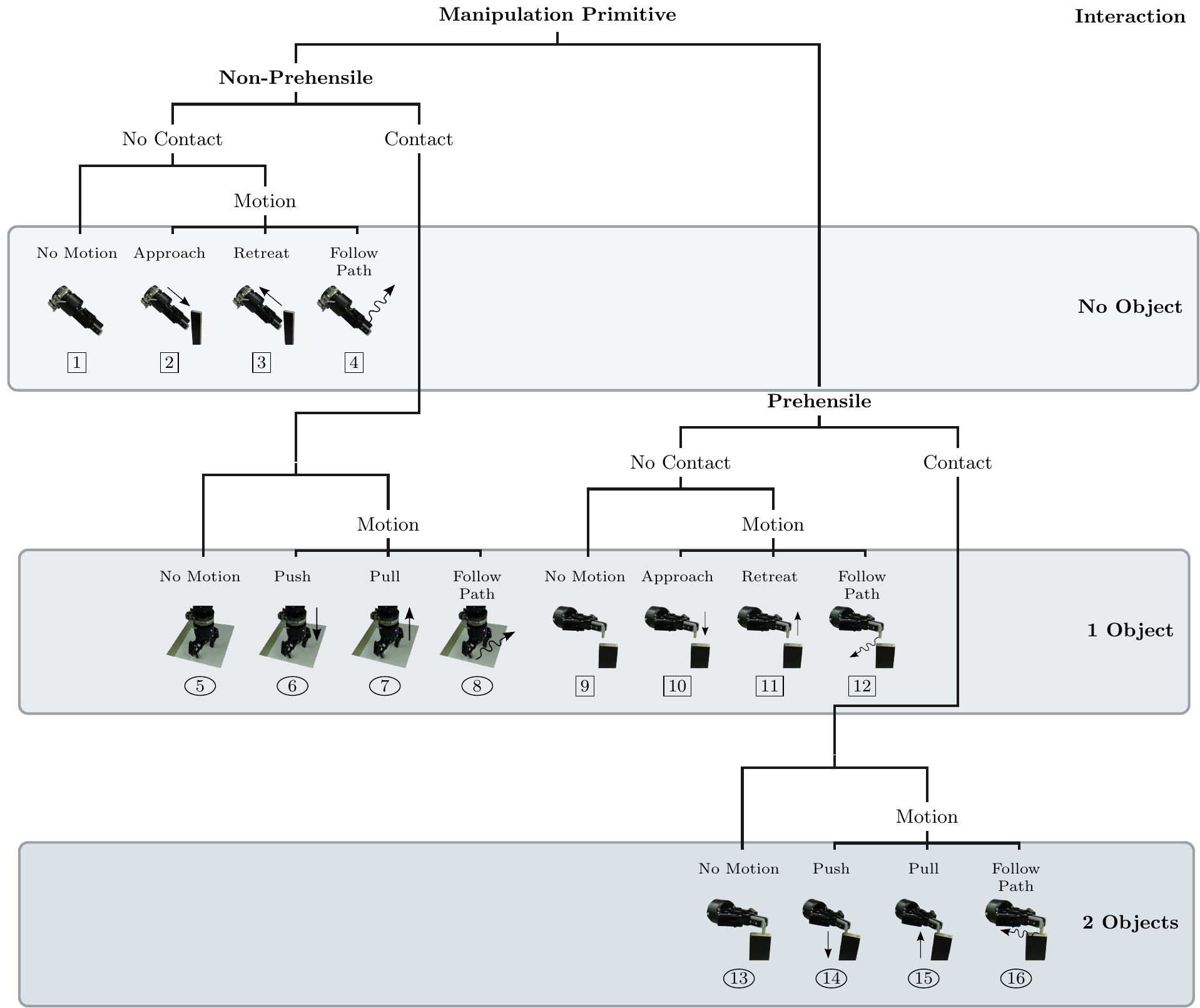}
  \caption{Manipulation taxonomy proposed in this work. Any robotic manipulation task can be classified using these primitives. There are three levels of difficulty depending on the number of objects the robot is interacting with. The control mode is indicated by the shape enclosing the primitive number. A rectangle indicates position-control mode. An ellipse indicates compliant-control mode.}
  \label{fig:taxonomy}
\end{figure*}
We present a motion-centric taxonomy that classifies manipulation primitives required for assembly tasks. Typically, two different approaches have been used in previous taxonomies: object- \cite{Cutkosky1990,Feix2009} or motion-centric \cite{Bullock2012,Owen-Hill2013}. Object-centric classifications define the primitives focusing on the characteristics of the manipulated object, which complicates their extension to different type of systems. On the other hand, motion-centric typologies allow to use different strategies to complete the same task. This flexible approach can be adapted to diverse manipulation systems depending on their capabilities. Our taxonomy focuses on industrial manipulators equipped with a parallel-jaw gripper. In the case of more complex end-effectors, in-hand manipulation primitives can be used \cite{Bullock2012}.


\subsection{Definitions}
\begin{itemize}
  \item \emph{Task}. High-level work to be done by the robotic manipulation system, e.g. bimanual insertion of a pin into a wood stick.
  \item \emph{Subtask}. Functional division of a task, e.g. grasping a pin, picking a stick, or inserting a pin. A task normally include several subtasks.
  \item \emph{Manipulation Primitive}. Basic action defined in the taxonomy. Typically, various primitives constitute a subtask. Moreover, they comprise the basic capabilities that the software framework provides for each manipulator. 
  \item \emph{Prehensile}. The hand/gripper can stabilize the object without need for external forces such as gravity. Basically, the object is grasped.
  \item \emph{Contact}. The hand/gripper or the object being grasped is touching any external body.
  \item \emph{Motion}. The end-effector is moving with respect to the robot's coordinate frame.
  \item \emph{Push/Pull}. A force is applied and the object being manipulated is moving as a result.
\end{itemize}

\subsection{Manipulation Taxonomy}
\label{sec:manipulation_taxonomy}
Taxonomies classify information into descriptive groups. In robotics, taxonomies are usually used to define the possible grasps of dexterous robotic hands~\cite{Cutkosky1990,Feix2009}. These kind of taxonomies focus mainly on in-hand movements and disregard the larger movements of a robotic manipulator.

Similarly to \cite{Bullock2012} and \cite{Owen-Hill2013}, a motion-centric approach has been adopted for the taxonomy proposed in this work.
It allows for greater flexibility than an object-centric approach, which would restrict the manipulation to the \emph{a~priori} knowledge of the object. The motion-centric approach is suitable for any manipulation performed by a hand-type manipulator.
\fref{fig:taxonomy} shows the motion-centric taxonomy proposed in this work. It is independent to the object being manipulated. For the classification of bimanual manipulation tasks, the primitives can describe the actions performed by each manipulator. 

\subsection{Primitives Requirements}
Once the manipulation primitives have been defined, it is needed to determine their specific requirements. As shown in \fref{fig:taxonomy}, there are three natural levels of difficulty depending on the number of objects involved in the manipulation. 
For a position-controlled manipulator, contact interactions represent an additional challenge, which prompts us to indicate the control mode required for each primitive in the taxonomy. 

\subsubsection{Position Mode}
This control mode is used for all the primitives that do not involve contact interactions. Despite of being inherently simple for a position-controlled manipulator, this mode requires precise localization of the objects when the robot moves in their proximity. For instance, if the robot needs to grasp and object, first it will approach to the grasp position (primitive \action{2}), but errors in the position estimation may result on the robot hitting the object and failing the task.

\subsubsection{Compliant Mode}
This mode is used for primitives where there are contact interactions between the gripper and the object or between the gripped object and another object. 
Depending on the task, force- or impedance-controlled motion will be used. One example of a force-controlled primitive is number \faction{5}. It can be used to maintain contact with a table while the gripper is closed to grasp a small object. In this case, controlling the force guarantees the contact between the gripper and the table and avoids unwanted interaction forces.
An example of an impedance-controlled primitive is number \faction{15}. Imagine a task where the force required to extract an object is unknown, therefore the force-controlled approach may fail.
Moreover, this compliant mode can be used to reduce the uncertainty in the localization of the objects. For instance, the robot can detect the exact position of the object once it detects a contact.

\section{Hardware and Software Platform}
\label{sec:setup}
The robotic platform used in this work is characterized by cost-efficient, off-the-shelf components combined with classical position-control industrial manipulators. This will help address the problems of fine assembly under unstructured environments at a limited cost.

The main components of the proposed platform are:
\begin{itemize}
  \item 2 $\times$ Denso VS060: Six-axis industrial manipulator.
  \item 2 $\times$ Robotiq Gripper 2-Finger 85: Parallel adaptive gripper designed for industrial applications. Closure position, velocity and force can be controlled. The gripper opening goes from $0$ to $85$ mm. The grip force ranges from $30$ to $100$ N.
  \item 2 $\times$ ATI Gamma \ac{ft} Sensor: It measures all six components of force and torque. They are calibrated with the following sensing ranges: $f = \left[32,32,100 \right]$ N and $\tau = \left[2.5,2.5,2.5 \right]$ Nm.
  \item 1 $\times$ Optitrack Motion Capture System: Six Prime 17W cameras that can track up to five rigid bodies. The error in position and orientation estimation is directly related to the amount of markers used per rigid body (minimum three markers are required). We have observed that the estimation error ranges between $\pm 0.5 - 3$ mm for the position and $\pm 0.01 - 0.05$ radians for the orientation. This estimation error is due to the diameter of the markers which ranges between $3$ to $10$ mm in our setup.
\end{itemize}

\begin{figure}[tb]
  \centering
  \includegraphics[width=0.51\linewidth ]{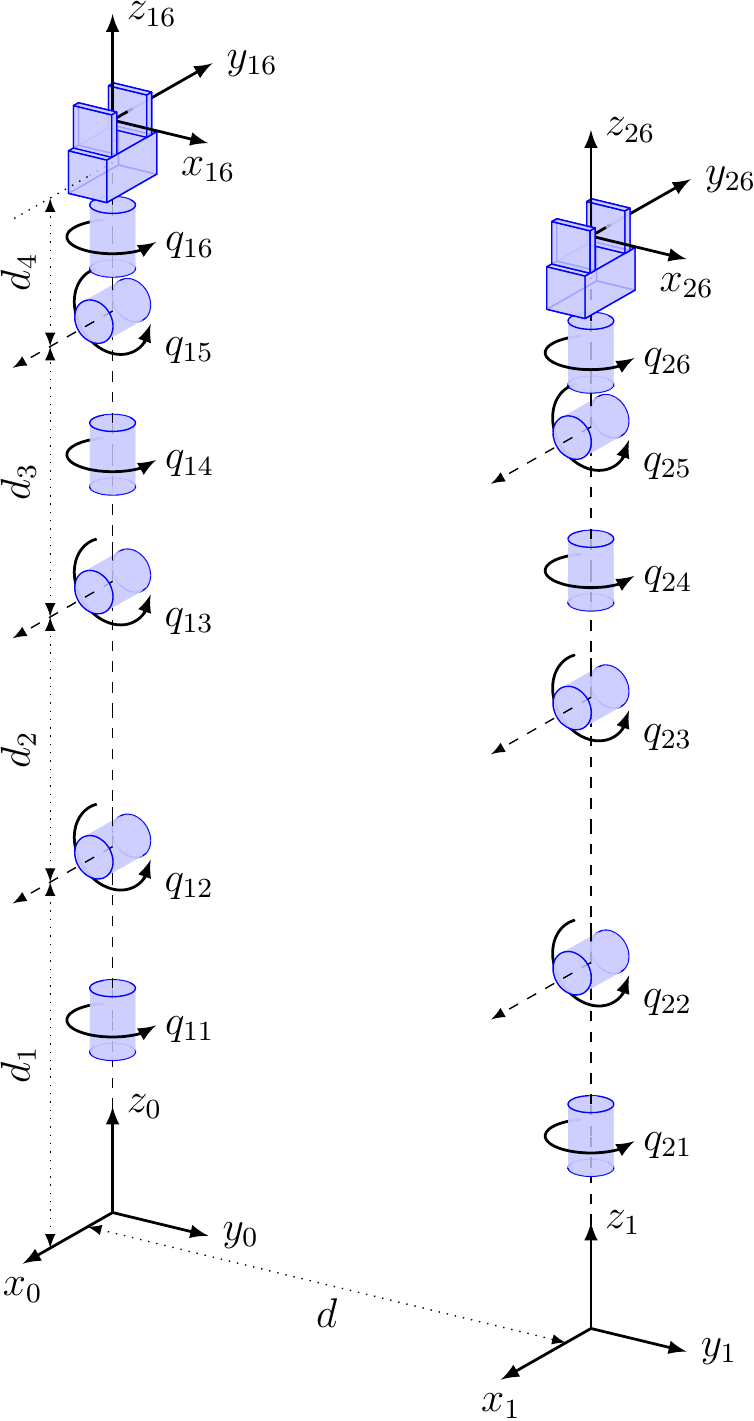}
  \caption{Kinematic diagram of the bimanual setup. The distance $d$ has been optimized to maximize the joint and intersection workspaces.}
  \label{fig:bimanual_kinematics}
\end{figure}

\subsection{Bimanual Workspace Optimization}
Appropriate values for the distance between the two robots ($d$ in \fref{fig:bimanual_kinematics}) can be selected either by trial and error or by solving an optimization problem with constraints imposed as a function of the resulting reachable workspace. The typical approach to quantify the manipulability of serial robots is to use a quality value for reachable positions along the robot's workspace. 
Normally the Yoshikawa's manipulability index \cite{Yoshikawa1985} is used:

\begin{equation}
w = \sqrt{\det{\left(\mat{J}\mat{J}^{T}\right)}}
\end{equation}

This value describes the distance to singular configurations but it does not consider the robot's joint limits. A modified index can be penalize to account for the effects of the joint limits on the manipulability of a serial manipulator. 

\begin{equation}
  P(\vect{q}) = \sum_{j=1}^{n}\dfrac{\left( l_{j}^{+} - l_{j}^{-} \right)^{2}}{4 \left( l_{j}^{+} - q_{j} \right)\left( q_{j} - l_{j}^{-} \right)} \, ,
\label{eq:penalization}
\end{equation}

We use the penalization function \eqref{eq:penalization} proposed in \cite{Dubey1995}, which results in the modified index $w^{*} = \frac{w}{P(\vect{q})}$.

Finally, we maximize a linear combination between the union and the intersection workspace that is a function of the distance $d$. The resulting workspace of the bimanual setup is shown in \fref{fig:bimanual_kinematics} where the optimized distance is $d = 1.042$ meters.

\begin{figure}[tb]
  \centering
  \includegraphics[width=0.9\linewidth ]{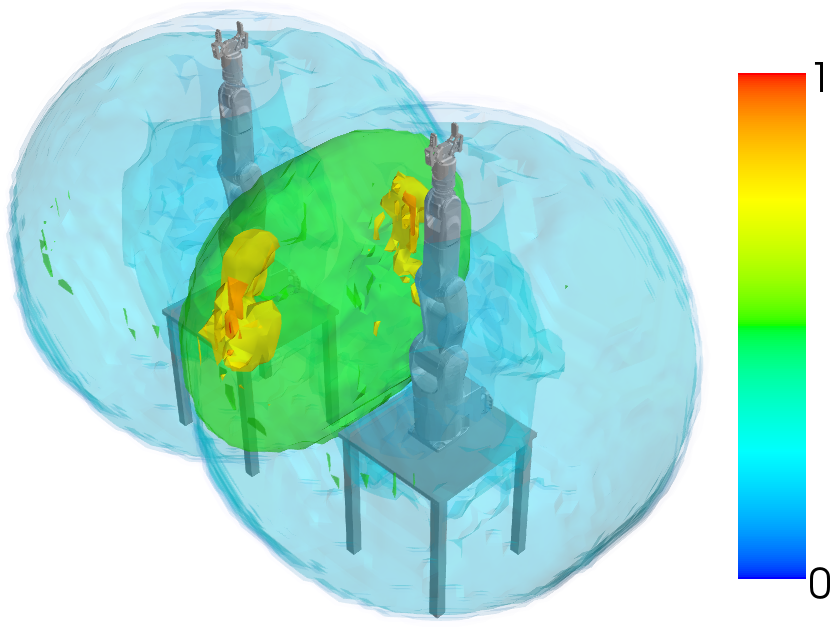}
  \caption{Normalized reachability of the bimanual setup. The workspace intersection shows the combined reachability of the two manipulators.}
  \label{fig:workspace}
\end{figure}

\subsection{Motion Planning in Free Space}
For all the collision-free movements the Bi-directional \acp{rrt} algorithm~\cite{Kuffner2000} available in OpenRAVE~\cite{diankov2010} is used. Similarly to LaValle in~\cite{Lavalle2006}, our system uses prioritized planning. The motion path is calculated for one robot at the time and is repeated until all the movements are completed. This reduces the overload in collision checking and avoids the need for path coordination between robots.

\subsection{External Wrenches Estimation}
In our setup, one \ac{ft} sensor is mounted at the wrist of each robot. It measures the dynamic effects of the end-effector and any external wrench due to contact interactions with the environment. External wrenches can be estimated by compensating the dynamic effects of the end-effector (weight and inertia) \cite{Swevers2007,Kubus2008,Hollerbach2008}. This approach requires the identification of the inertial parameter of the end-effector. We propose an off-line approach which only uses the \ac{ft} sensor measurements along a defined trajectory.

\subsubsection{Optimal Excitation Trajectories}
During the identification process, it is necessary to ensure that the excitation is sufficient to provide accurate and fast parameter estimation in the presence of disturbances, and that the collected data is simple and yields reliable results.
First, a trajectory parametrization is selected, and second the trajectory parameters are calculated by means of optimization.

The excitation trajectory for each joint has been chosen as a finite sum of harmonic sine and cosine functions, similar to \cite{Swevers2007,Kubus2008}. Each one with a total of $2N + 1$ parameters, which correspond to the degrees of freedom of the optimization problem.
\begin{align}
  q_{j}\left(t\right) &= \sum_{k=1}^{N}{\dfrac{a_{j}^{k}}{w_{f}k}\sin\left( w_{f}kt \right) - \dfrac{b_{j}^{k}}{{w_{f}k}}\cos\left( w_{f}kt \right)} + q_{j}^{0} \\
  \dot{q}_{j}\left(t\right) &= \sum_{k=1}^{N}{a_{j}^{k}\cos\left( w_{f}kt \right) + b_{j}^{k}\sin\left( w_{f}kt \right)} \\
  \ddot{q}_{j}\left(t\right) &= \sum_{k=1}^{N}{-a_{j}^{k}w_{f}k\sin\left( w_{f}kt \right) + b_{j}^{k}w_{f}k\cos\left( w_{f}kt \right)} \, .
\end{align}
The coefficients $a_{j}^{k}$ and $b_{j}^{k}$ are the amplitudes of the sine and cosine functions. $q_{j}^{0}$ is the offset of the position trajectory.

\begin{figure}[tb]
  \centering
  \includegraphics[width=0.9\linewidth ]{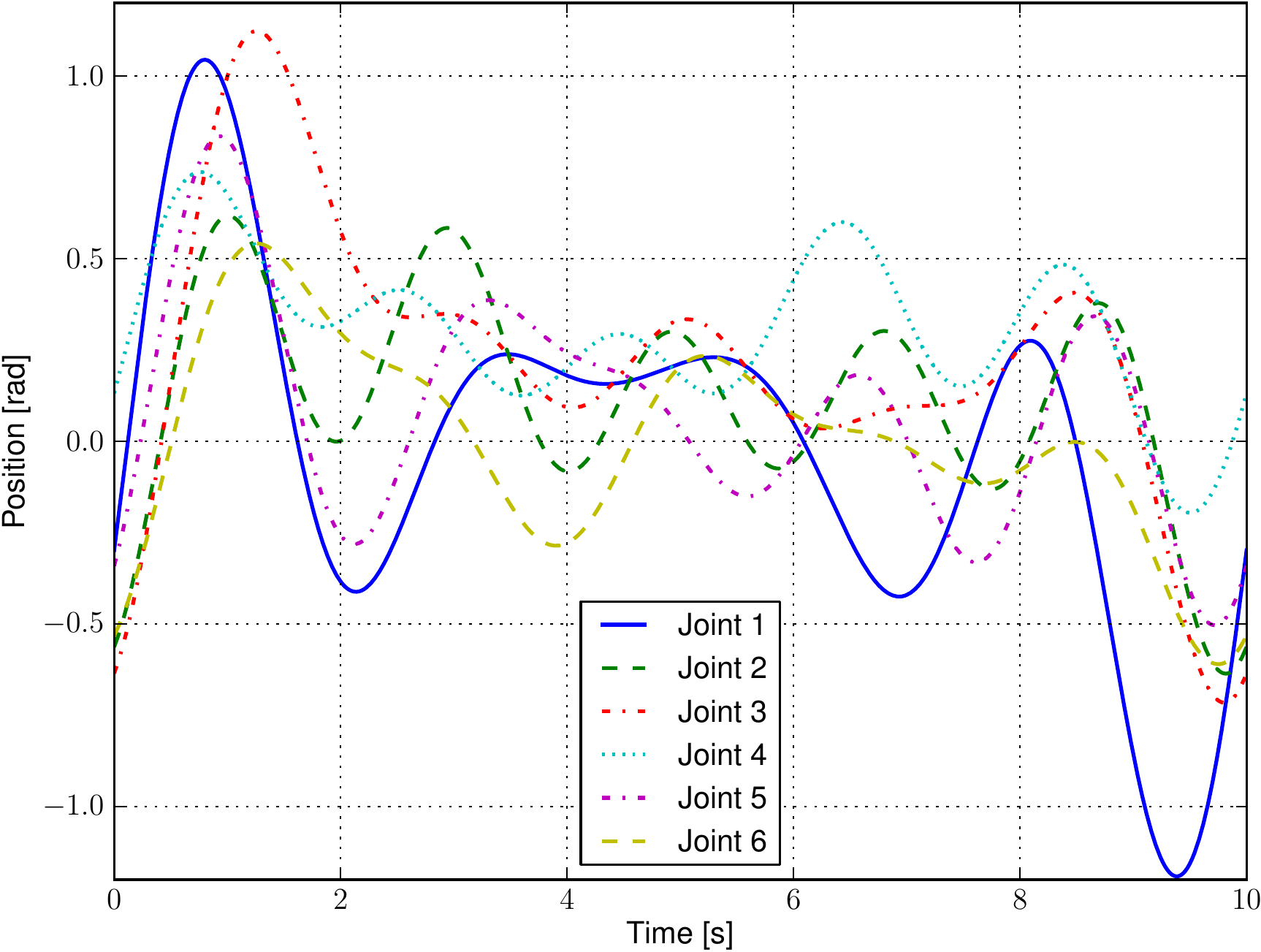}
  \caption{Optimized robot-excitation trajectory. One period of the optimized joint trajectories is shown. These trajectories consist of a five-term Fourier series with a base frequency of $0.1$ Hz. The trajectory parameters are optimized according to the $d$-optimality criterion, taking into account workspace limitations, and constraints on joint velocities and accelerations.}
  \label{fig:excitation_trajectories}
\end{figure}

\fref{fig:excitation_trajectories} shows the optimized trajectory for the six robot's joints ($11$ parameters per joint). The base frequency has been selected in order to cover a larger part of the robot workspace for the given maximum joint velocities and accelerations, even thought it requires a longer measurement time. The identification process is performed off-line for each end-effector.

\subsubsection{End-effector Dynamics}
The wrist-mounted \ac{ft} sensor is measuring the loads on the last link excluding itself. In particular, since the end-effector is always present, it is possible to compensate the wrench it generates by determining its inertial parameters. The Newton-Euler equation of this last body refereed to the \ac{ft} sensor frame $O_{s}$ is,

\begin{equation}
  \vect{f}_{s} = \mat{I}_{s}\vect{a}_{s} + \vect{v}_{s}\times\mat{I}_{s}\vect{v}_{s} \, ,
\end{equation}

where the resulting spatial force $\vect{f}_{s}$ is a function of the spatial inertia $\mat{I}_{s}$, the spatial acceleration $\vect{a}_{s}$, and the spatial velocity $\vect{v}_{s}$.
As shown in \cite{Hollerbach2008}, the force and torque measurements by the wrist sensor must be expressed in terms of the product of known values and the unknown inertial parameters. The measured wrench $\vect{f}_{s}$ can be written as:

\begin{equation}
\label{eq:fs_estimation}
  \resizebox{0.9\hsize}{!}{$
  \vect{f}_{s} =
  \begin{bmatrix}
    \vect{a}_{s} & \matop{S}{\dot{\vect{\omega}}_{s}} + \matop{S}{\vect{\omega}_{s}}\matop{S}{\vect{\omega}_{s}} & \mat{0} \\
    \mat{0} & -\matop{S}{\vect{a}_{s}} & \matop{L}{\dot{\vect{\omega}}_{s}} + \matop{S}{\vect{\omega}_{s}}\matop{L}{\vect{\omega}_{s}} \\
  \end{bmatrix}
  \begin{bmatrix}
  m_{s}         \\
  \vect{c}_{s} \\
  \matop{l}{\mat{I}_{s}} \\
  \end{bmatrix} $}
\end{equation}

where $\matop{L}{\vect{\omega}_{s}}$ is a $3 \times 6$ matrix of angular velocity elements, $\matop{l}{\mat{I}_{s}}$ is the inertia matrix vectorized and $\matop{S}{\vect{\omega}_{s}}$ is the skew-symmetric matrix.
\eqref{eq:fs_estimation} can be expressed more compactly as,

\begin{equation}
  \vect{f}_{s} = \mat{A}_{s}\vect{\phi}_{s} \, ,
\end{equation}

where $\mat{A}_{s}$ is a $6 \times 10$ matrix, and $\vect{\phi}_{s}$ is the vector of the 10 unknown inertial parameters.
To estimate the force/torque offsets, the matrix $\mat{A}_{s}$ is augmented by the identity matrix $\mat{E}$ and the parameters vector $\vect{\phi}_{s}$ is expanded to include the offsets $\vect{f}_{0}$ and $\vect{\tau}_{0}$ to be estimated:

\begin{align*}
  \vect{\phi}_{s}^{*} &= 
  \begin{bmatrix}
    \vect{f}_{0} & \vect{\tau}_{0} & m_{s} & \vect{c}_{s} & \matop{l}{\mat{I}_{s}}
  \end{bmatrix}^{T} \\
  \mat{A}_{s}^{*} &= 
  \begin{bmatrix}
    \mat{E}_{6 \times 6} & \mat{A}_{s} \\
  \end{bmatrix}
\end{align*}

\subsection{Position-Based Explicit Force Control}
\begin{figure}[tb]
  \centering
  \includegraphics[width=\linewidth ]{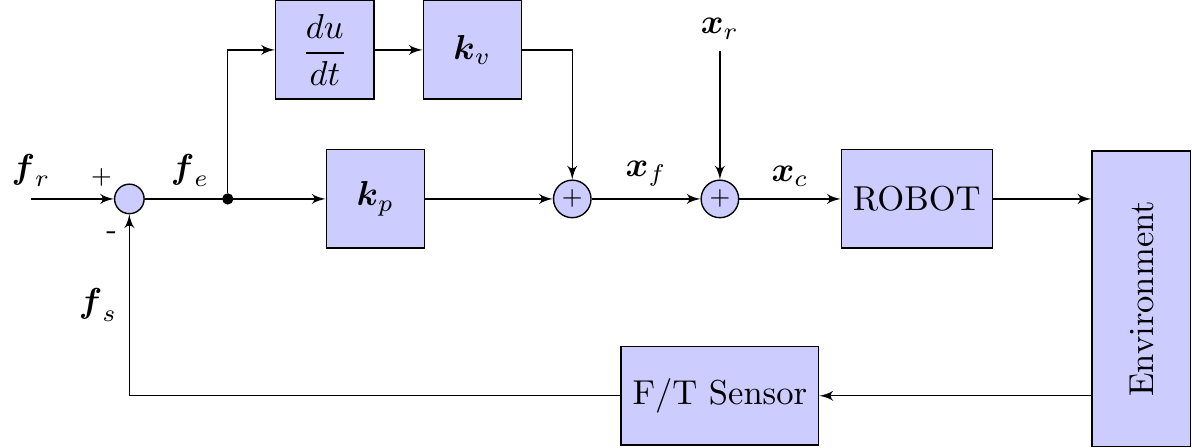}
  \caption{Position-based explicit force control.}
  \label{fig:explicit_force_control}
\end{figure}

The idea of a position-based explicit controller is to take a position-controlled manipulator as a baseline system and make the necessary modifications to achieve compliant motion control~\cite{Ott2010,Seraji1994}.
\fref{fig:explicit_force_control} shows the adopted force control scheme, where $\vect{x}_{r}$ is the reference position and $\vect{f}_{r}$ the force setpoint when the robots interacts with the environment. The contact force $\vect{f}_{s}$ is fed back to the force compensator which produces a perturbation $\vect{x}_{f}$, so that the end-effector tracks the modified commanded trajectory $\vect{x}_{c}$. Thus the force feedback law is given by
\begin{equation}
  \vect{x}_{f}(t) = \vect{k}_{p}\vect{f}_{e}(t) + \vect{k}_{v}\dot{\vect{f}_{e}}(t)\, .
  \label{eq:force_control}
\end{equation}
This controller ensures uniform performance when in contact with environments having unknown stiffness. For details regarding the controller's robustness see \cite{Seraji1994,Calanca2015}.

Currently, we have tested two compliant controllers: explicit force control and admittance control. Initially, our idea was to implement the explicit force controller only for the contact-without-motion primitives (\faction{5} and \faction{13}) but after some experimental trials, we found that this controller was also capable of performing primitives \faction{14} and \faction{16}. Therefore, for the bimanual pin insertion task, we use the explicit force controller for the compliant mode. We believe that for the next steps of our long term project (assembling an IKEA chair), more complex compliant controllers will be needed, specially for bimanual collaborative manipulation tasks like flipping the chair using both arms.

\section{Example: Bimanual Pin Insertion}
\label{sec:task}
As discussed in \sref{sec:introduction}, we have chosen a bimanual pin insertion task for the evaluation of the proposed framework.
This task starts with a cylindrical pin ($r=4$~mm, $l=30$~mm) and a wood stick ($20\times 50 \times 270$ mm) on a table. The left arm picks the pin, the right arm picks the stick, and both arms move to the insertion area (a location where the two manipulators can reach). The left arm uses the pin to explore the stick and to find the hole. Once it finds the hole, inserts and releases the pin.

The task is naturally divided into three mid-level sub-tasks:
\begin{itemize}
  \item Compliant grasp of the pin,
  \item Pick \& place the stick, and
  \item Compliant insertion of the pin
\end{itemize}

\begin{figure*}[t]
  \centering
  \includegraphics[width=\linewidth ]{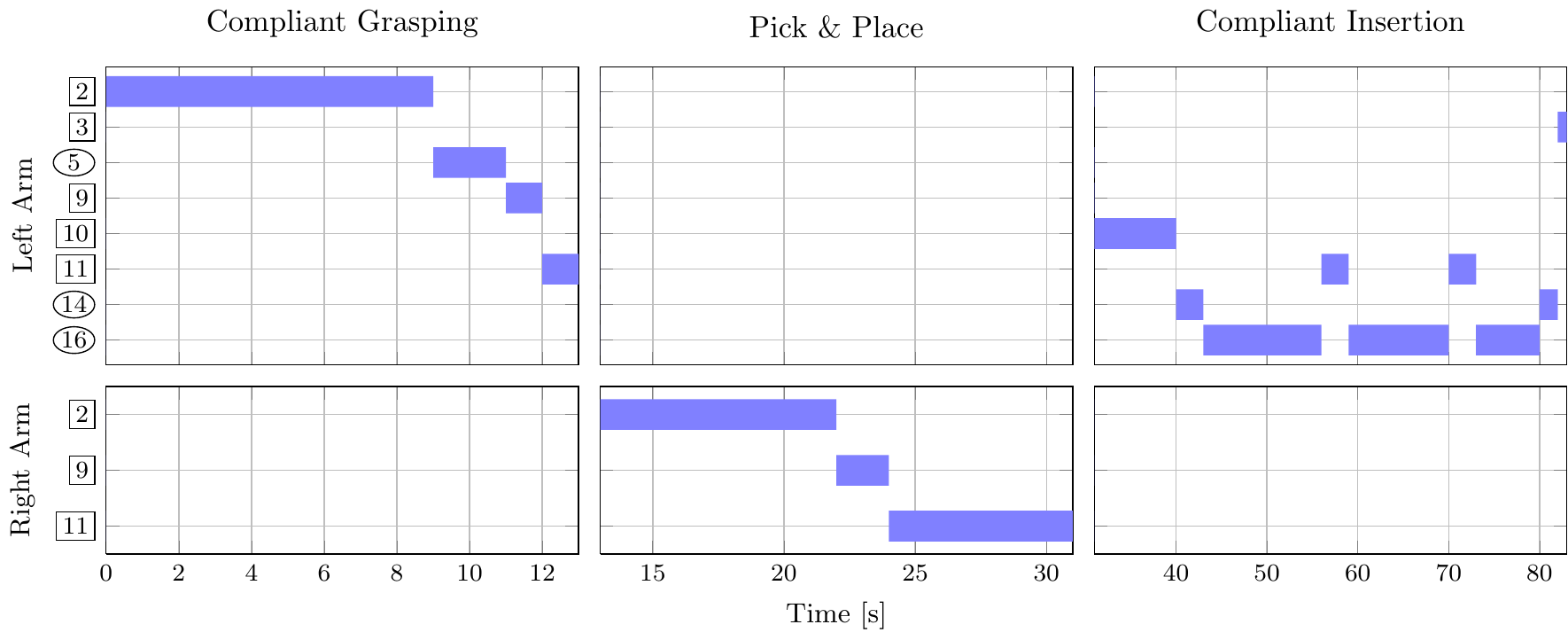}
  \vspace*{-8mm}
  \caption{Time-line representation of the manipulation primitives used for the bimanual pin insertion task. Only six (6) manipulation primitives are required.}
  \label{fig:task_transitions}
\end{figure*}
\begin{figure*}[t]
  \centering
  \subfloat[$t=0$s.]{\includegraphics[height=30mm]{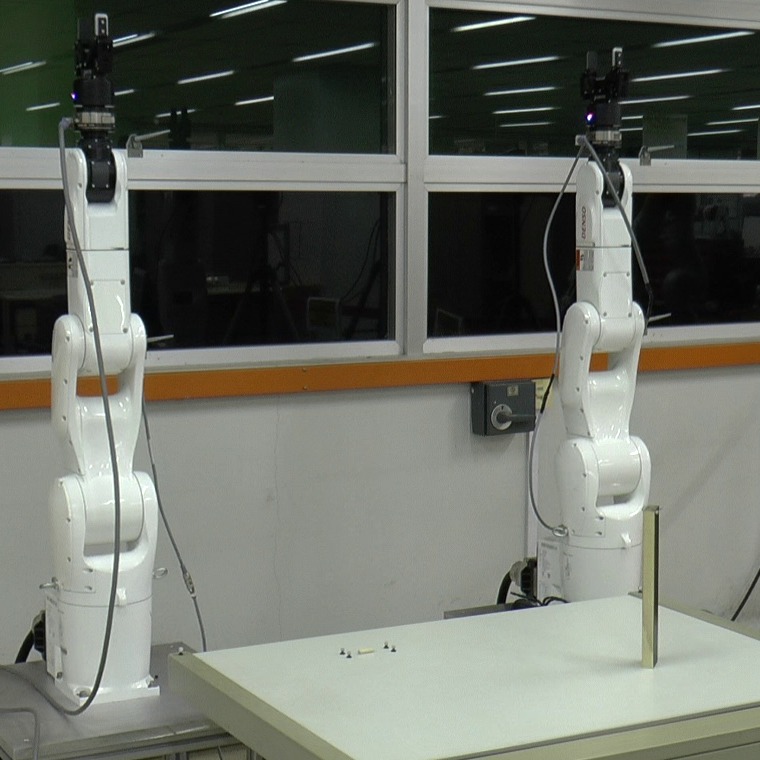}}\quad
  \subfloat[$t=11$s.]{\includegraphics[height=30mm]{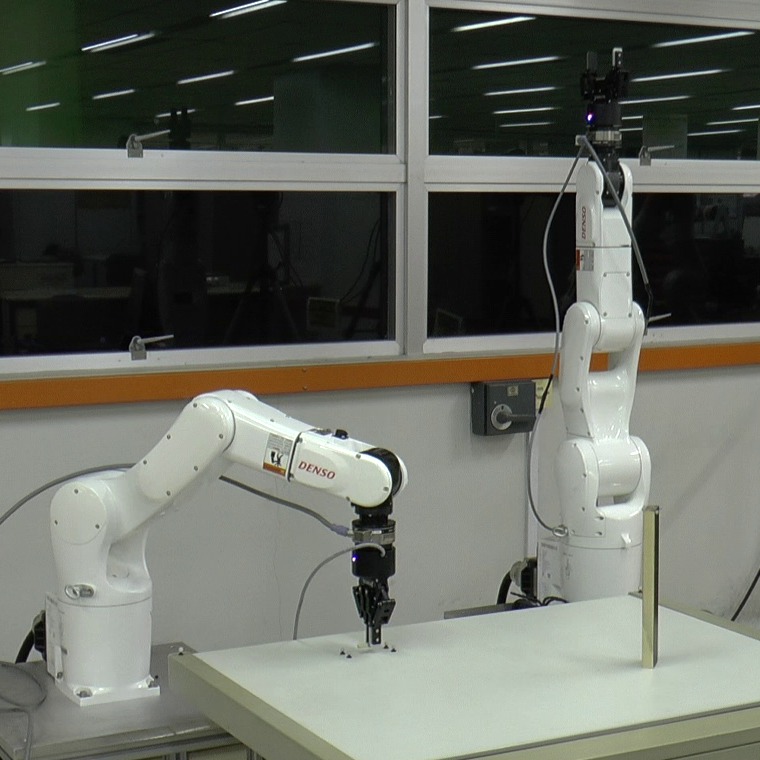}}\quad
  \subfloat[$t=24$s.]{\includegraphics[height=30mm]{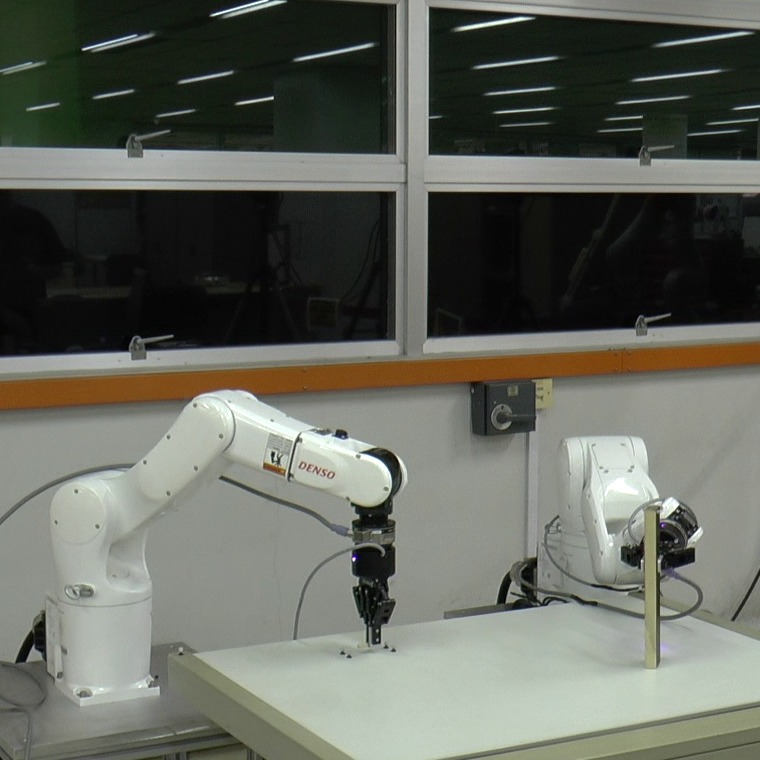}}\quad
  \subfloat[$t=31$s.]{\includegraphics[height=30mm]{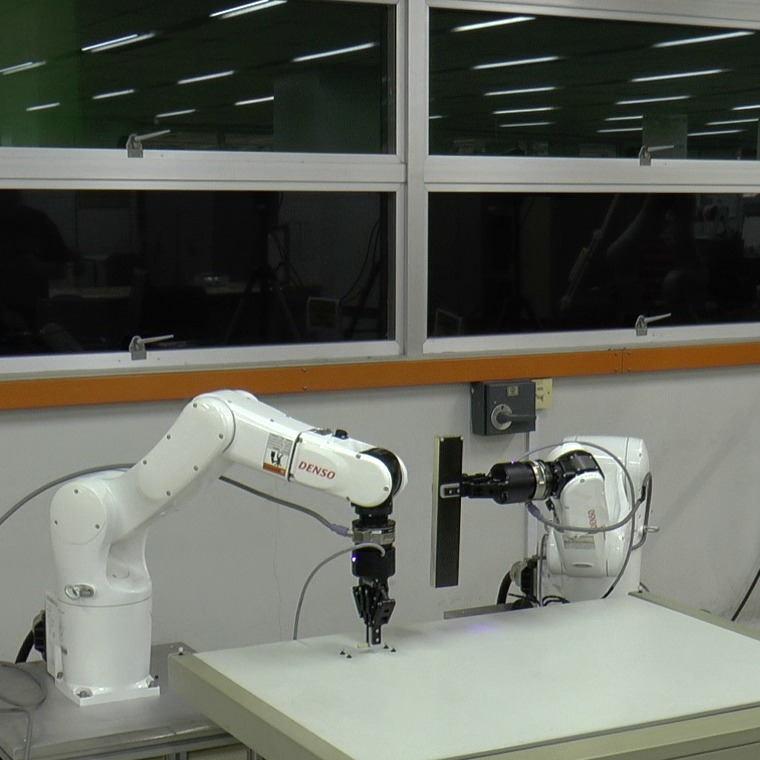}}\quad
  \subfloat[$t=43$s.]{\includegraphics[height=30mm]{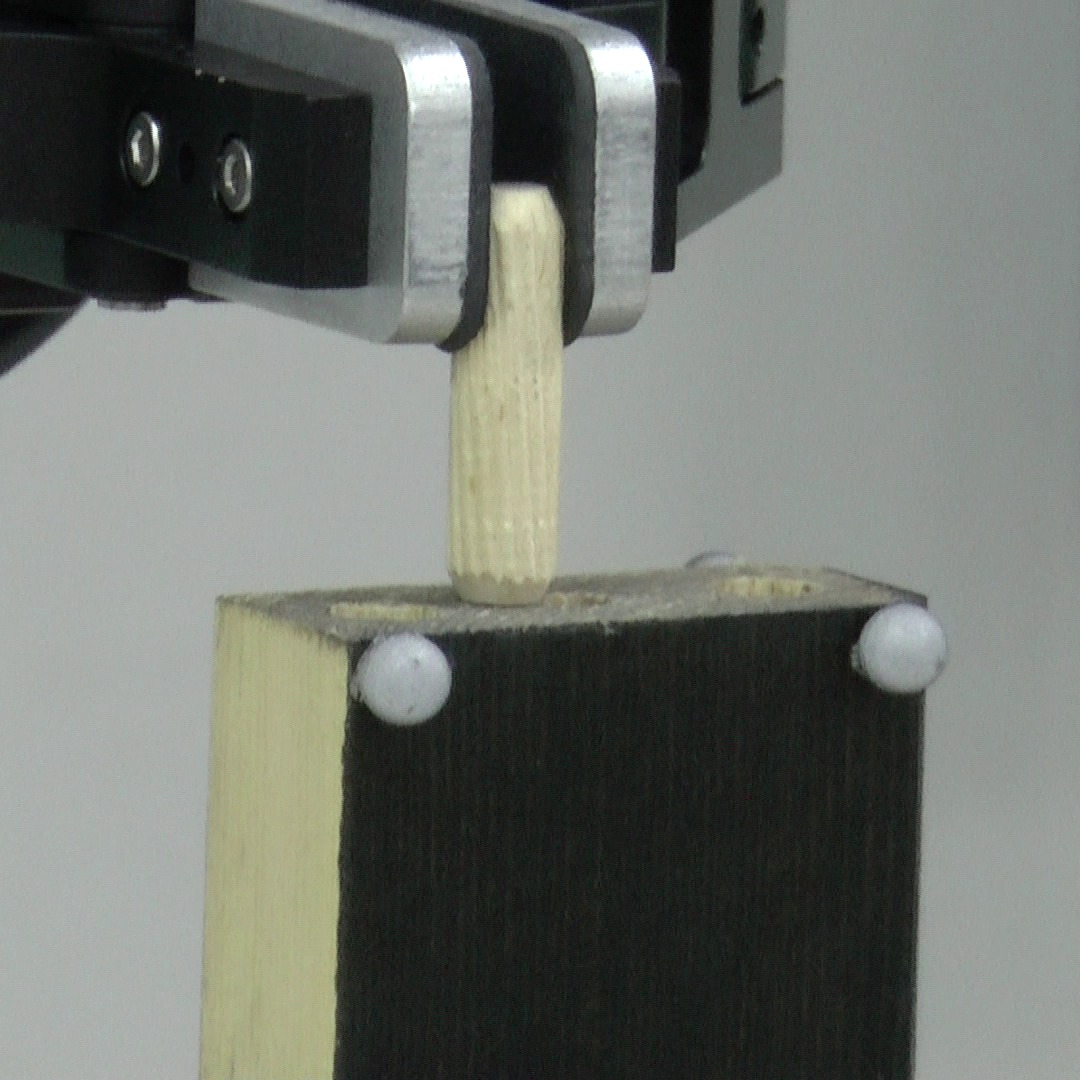}} \\
  \subfloat[$t=56$s.]{\includegraphics[height=30mm]{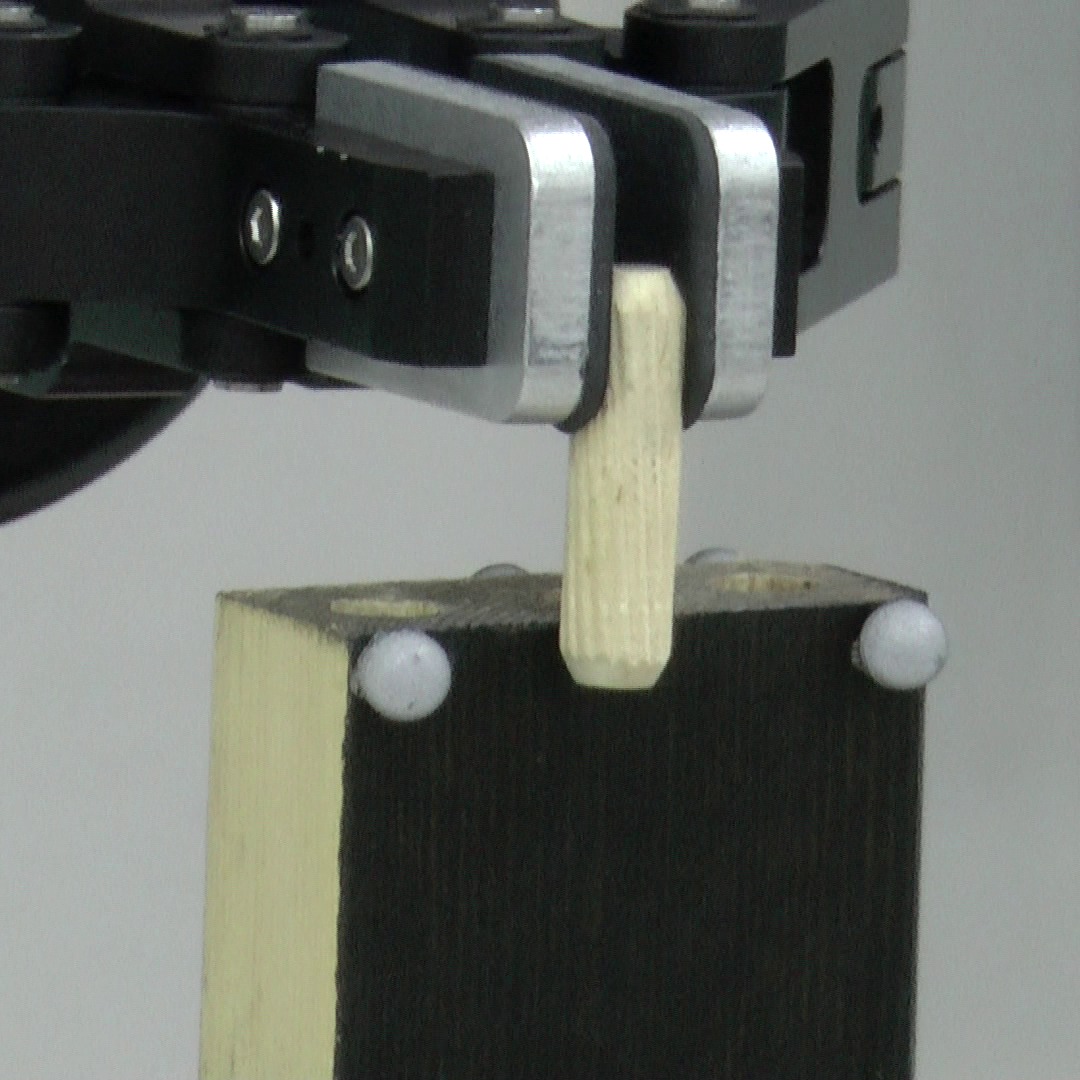}}\quad
  \subfloat[$t=70$s.]{\includegraphics[height=30mm]{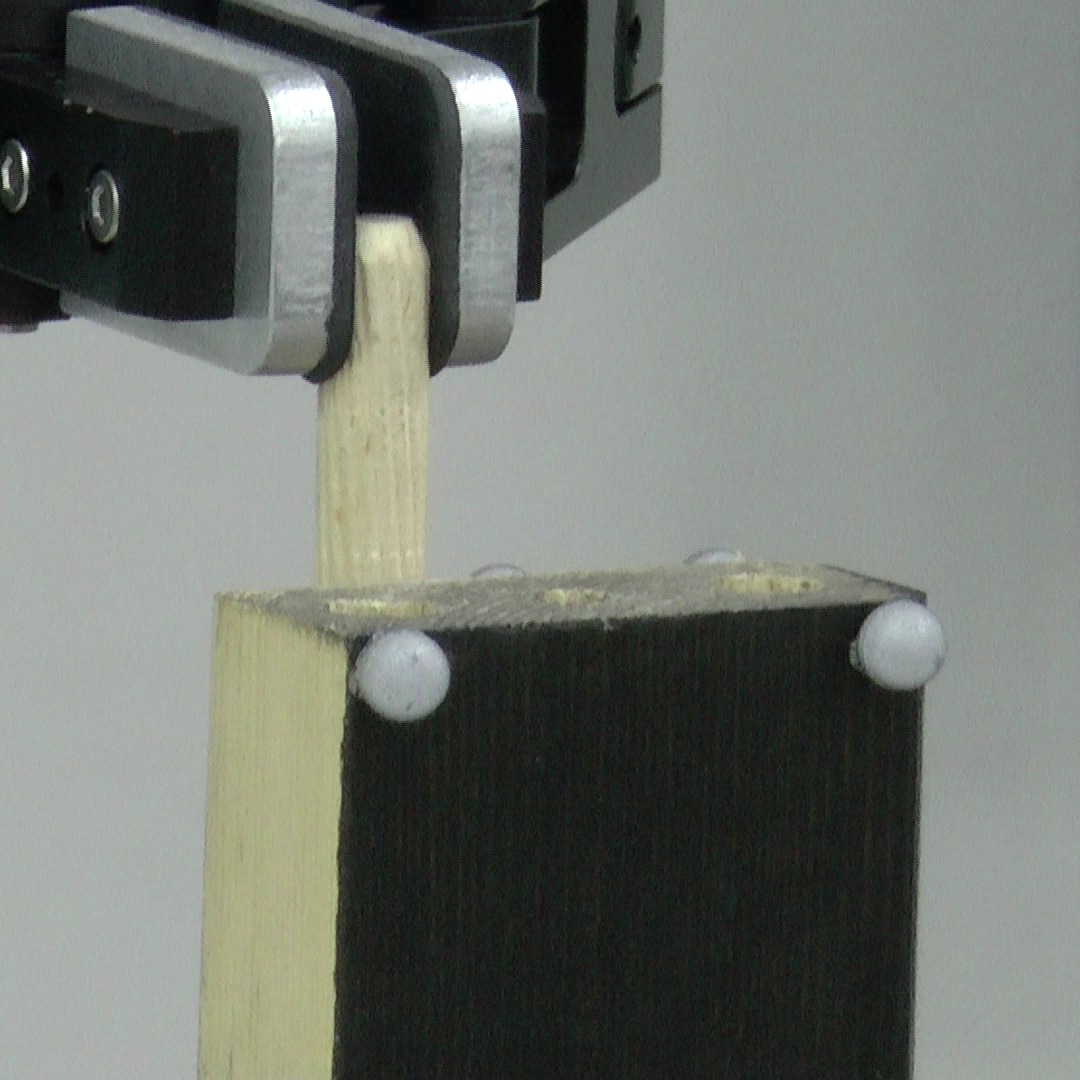}}\quad
  \subfloat[$t=80$s.]{\includegraphics[height=30mm]{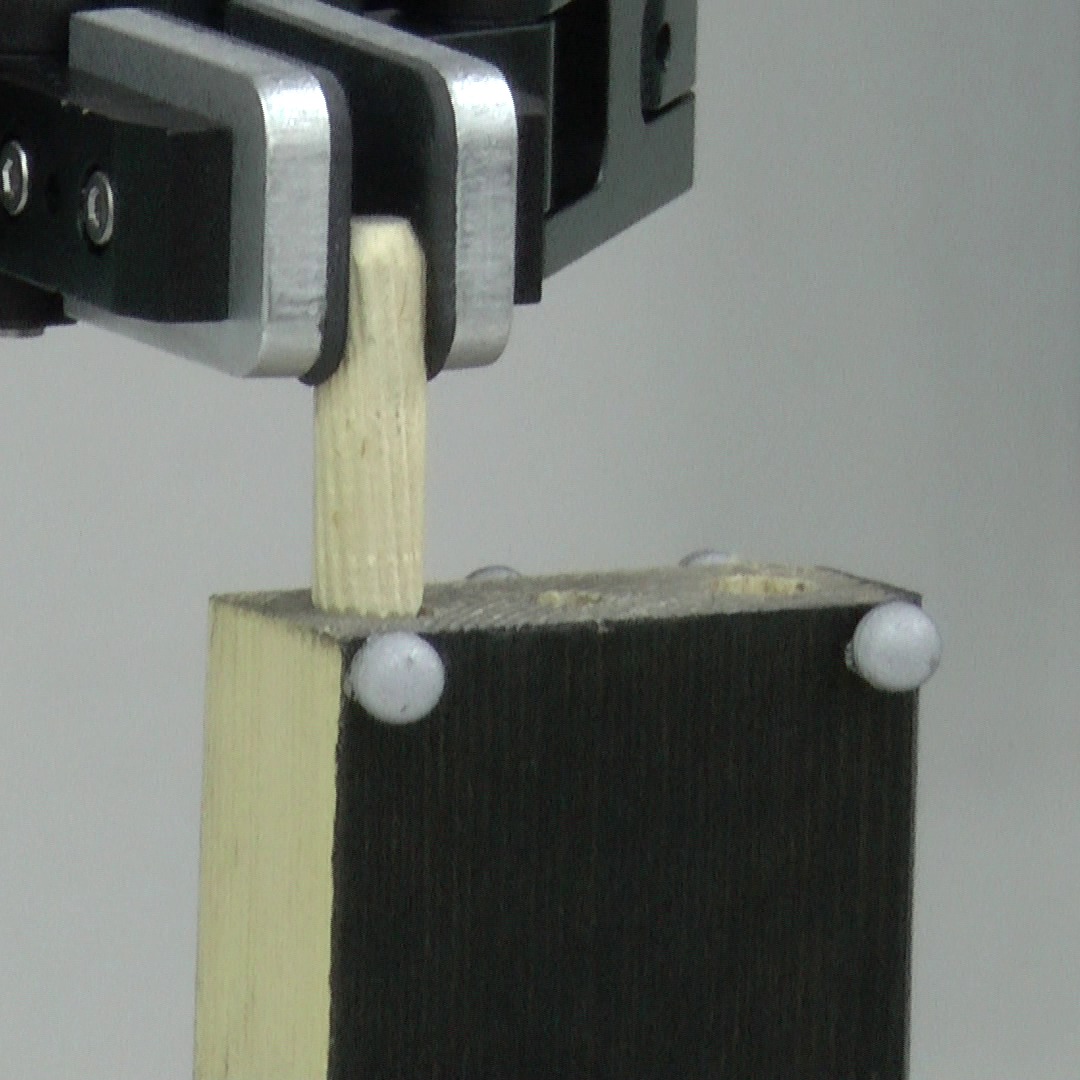}}\quad
  \subfloat[$t=82$s.]{\includegraphics[height=30mm]{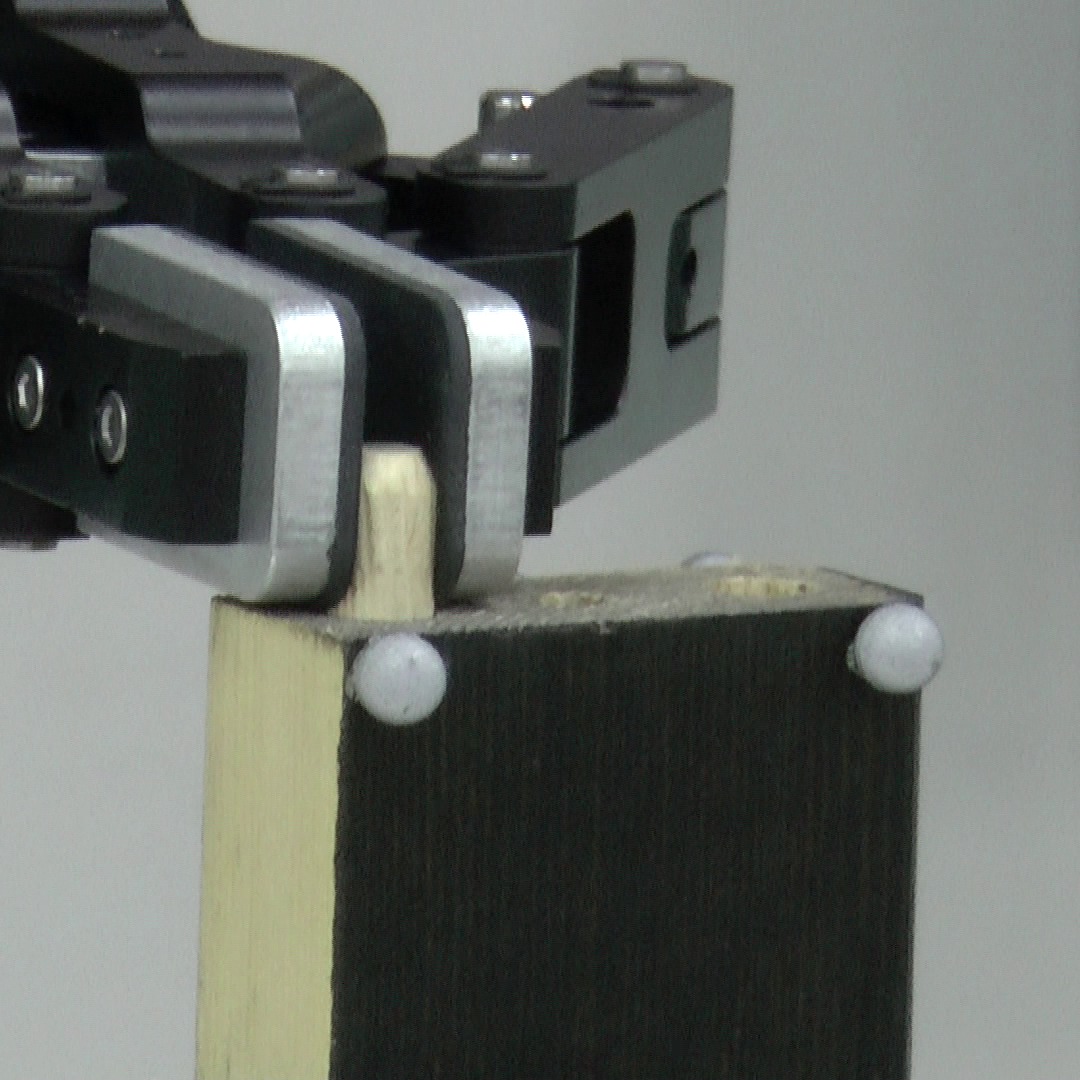}}\quad
  \subfloat[$t=83$s.]{\includegraphics[height=30mm]{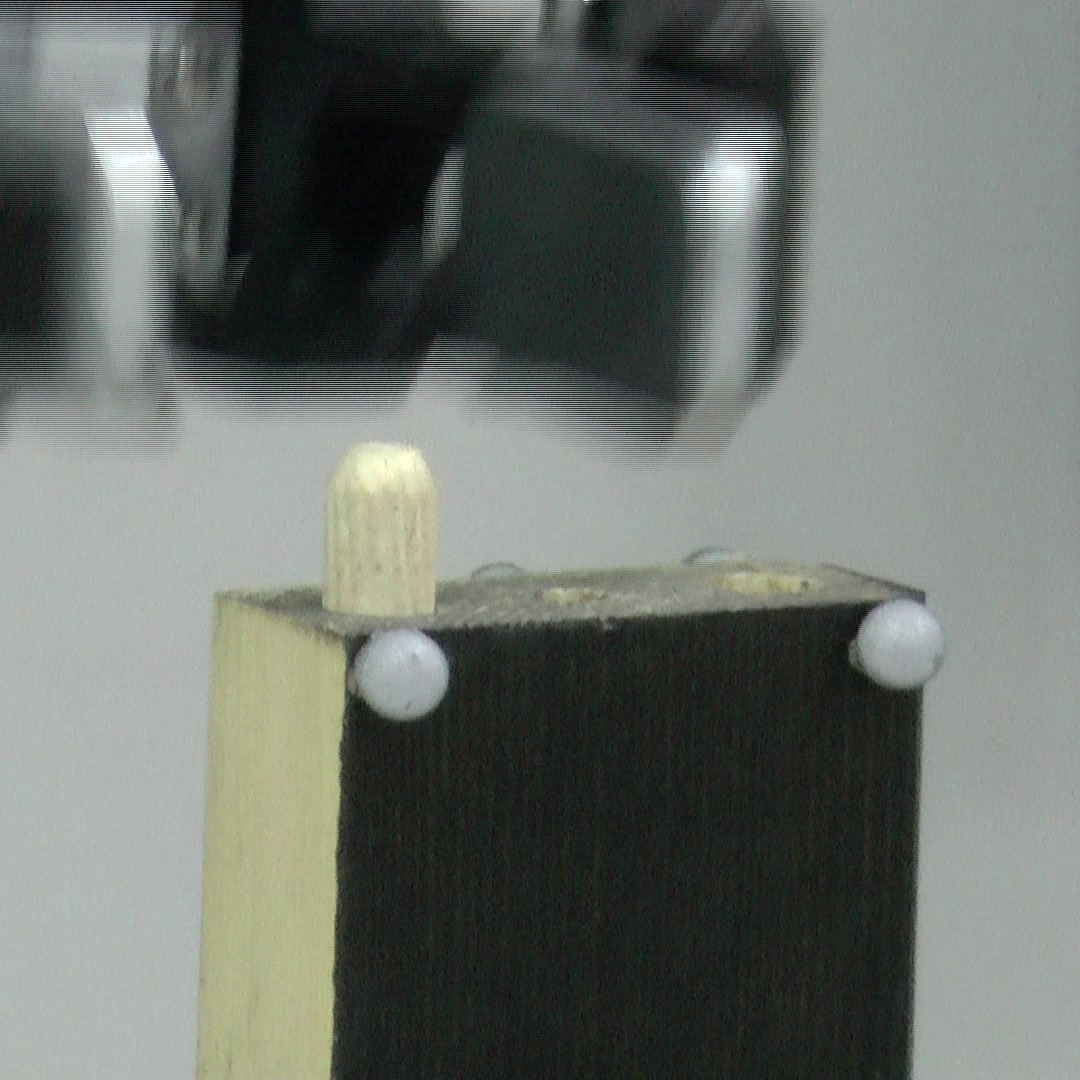}} \\
  \caption{Snapshots of the bimanual pin insertion. a) The initial position. The positions of the table, stick and pin are determined using Optitrack. b) The left arm performs the compliant grasping of the pin. c) The right arm grasps the stick. d) The right arm \enquote{places} the stick in a position where the insertion can take place. e) The left arm moves above the stick and detects the contact with the pin. f) Through force exploration, the left arm finds the first edge of the stick. g) The left arm finds the second edge of the stick. h) Using the refined position of the two edges, the system knows where the middle axis is and can find the hole. i) Once the hole is found, the left arm inserts the pin. j) Finally, the left arm releases the pin and moves back to its home position. The complete video can be found at \protect\url{http://goo.gl/cYl9sq}.}
  \label{fig:snapshots}
\end{figure*}

\subsection{Compliant Grasping}
Moving the left arm to grasp directly the pin is not possible due to two uncertainties in the system: First, the position error in the perception system can be up to $\pm 3$ mm. Second, the difference in the height of the gripper between the opened and the closed position is $13.9$ mm.
These two factors along with the mechanical compliance of the gripper at the tip (intended for encompassing grip) require extra capabilities to grasp small objects from the table top.

To cope with this, we use a compliant grasping approach. Initially, the robot moves to a pregrasp position, just above the pin, then it moves down until it detects contact with the table. The position-based explicit force controller is then used to maintain the contact with the table within a safe value that does not overcome the compliance of the gripper's tip. Finally the gripper is closed while the force controller maintains the contact with the table.

\subsection{Pick \& Place}
This is a simpler sub-task. The right arm picks the stick and \enquote{place} it in the insertion area. It needs to be hold tightly so that the left arm can perform the exploration, find the hole and insert the pin.

\begin{table}[tb]
  \caption{Parameters of the peg-in-hole task used for the assessment of the bimanual manipulation system.}
  \label{tab:peg_in_hole}
  \centering
  \begin{tabular}{l c c}
    \toprule
    \multicolumn{1}{c}{\textbf{Parameter}}   & \textbf{Symbol}    & \textbf{Value}\\
    \midrule
    Hole diameter   & $d_H$   & $8.1$ mm   \\
    Peg diameter    & $d_P$   & $8$ mm   \\
    Peg height      & $h$     & $30$ mm    \\
    \bottomrule
  \end{tabular}
\end{table}
\begin{table}[tb]
  \caption{Manipulation primitives used for the bimanual pin insertion task. The time to task completion is 83 seconds.}
  \label{tab:task_primitives}
  \centering
  \begin{tabular}{l l c l}
    \toprule
    \multicolumn{2}{c}{\textbf{Time} [s]} & \multirow{2}{*}{\textbf{Primitive}} & \multicolumn{1}{c}{\multirow{2}{*}{\textbf{Action}}}\\
    \multicolumn{1}{c}{\textbf{Start}} &  \multicolumn{1}{c}{\textbf{End}} & & \\
    \midrule
    \multicolumn{4}{l}{\emph{Compliant Grasping (left arm, 13 seconds.)}} \\
    $0$   & $9$   & \action{2}  & Approach to the pregrasp position  \\
    $9$   & $10$  & \faction{5}  & Contact with the table  \\
    $10$  & $11$  & \faction{5}  & Close gripper maintaining contact \\
    $11$  & $12$  & \action{9} & Grasp the pin \\
    $12$  & $13$  & \action{11} & Pick-up the pin from the table \\[1mm]
    \multicolumn{4}{l}{\emph{Pick \& Place (right arm, 18 seconds.)}} \\
    $13$  & $22$  & \action{2}  & Approach to the grasp position  \\
    $22$  & $24$  & \action{9} & Grasp the stick  \\
    $24$  & $31$  & \action{11} & Move the stick to the insertion area \\[1mm]
    \multicolumn{4}{l}{\emph{Compliant Pin Insertion (left arm, 52 seconds.)}} \\
    $31$  & $40$  & \action{10} & Move the pin above the stick  \\
    $40$  & $43$  & \faction{14} & Contact between the pin and stick  \\
    $43$  & $56$  & \faction{16} & Detect first edge of the stick  \\
    $56$  & $59$  & \action{11} & Move above and contact the stick  \\
    $59$  & $70$  & \faction{16} & Detect second edge of the stick  \\
    $70$  & $73$  & \action{11} & Move above and contact the stick  \\
    $73$  & $80$  & \faction{16} & Find the hole  \\
    $80$  & $82$  & \faction{14} & Insert the pin  \\
    $82$  & $83$  & \action{3}  & Release the pin  \\
    \bottomrule
  \end{tabular}
\end{table}
\subsection{Compliant Pin Insertion}
\label{sub:compliant_insertion}
The insertion sub-task is of the peg-in-hole type. This kind of setup is generally characterized using a precision value defined as,
\begin{equation}
  I = \log{}_2 \left( \dfrac{d_H}{d_H-d_P} \right) \, ,
  \label{eq:bax_peg_in_hole}
\end{equation}
where $d_H$ is the diameter of the hole and $d_P$ is the diameter of the peg. \tref{tab:peg_in_hole} shows the parameters for our pin insertion setup, which has a precision value $I=6.34$ bits. Other studies have used precision values within the same order of magnitude \cite{Hannaford1991,Yip2011} in telemanipulation applications where the operator deals with the challenge of localizing precisely the hole.

Due to the uncertainties on the position of the objects (pin and stick), the exact position of the holes is unknown. Moreover, given the parameters of the peg-in-hole setup (\tref{tab:peg_in_hole}), we have observed that the insertion fails for position errors above $0.5$ mm. To cope with these problems, we perform a force-controlled exploration of the wood stick using the pin.
The left arm moves above the stick with the pin grasped, then starts moving down until contact is detected. Next, we look for two edges of the stick. Considering that its dimensions are known, after finding the edges, the middle axis of the holes can be calculated. The robot \enquote{scratches} the pin over the stick following that axis until it finds the hole. After that, a force-controlled motion is carried out $\vect{f}=\left[ 0,0,-f_{z}\right]$ to insert the pin. This value ensures that the pin will move only in the $-\vect{z}$ direction until it reaches the bottom of the hole or the gripper touches the stick. 
From \eqref{eq:force_control}, it can be seen that the pin is driven down the distance $\vect{x}_{f}$ until the force error $\vect{f}_{e}$ equals to zero. This motion emulates and spring-damper impedance that depends on the compensator gains $\vect{k}_{p}$ and $\vect{k}_{v}$.
Finally the left arm releases the pin and moves back to its home position.

\tref{tab:task_primitives} depicts the manipulation primitives required for each sub-task with their corresponding times. The time to task completion for the bimanual pin insertion is 83 seconds. \fref{fig:task_transitions} depicts the transitions between manipulation primitives in a time-line representation.
\fref{fig:snapshots} shows snapshots of the bimanual pin insertion where each sub-task can be visually identify. The complete video can be found at \url{http://goo.gl/cYl9sq}.

\section{Conclusions and Future Work}
\label{sec:conclusions}
This paper introduced a complete framework for fine assembly tasks using industrial robots. We have presented a new taxonomy of manipulation primitives tailored for industrial fine assembly. This taxonomy focuses on parallel-jaw grippers and interactions with single or multiple objects which are the essence of assembly tasks.
Moreover, we have discussed the development and implementation of a software and hardware framework for bimanual manipulation. Our experimental setup shows that fine assembly manipulation can be successfully implemented on an industrial system that was originally built to be position-controlled. 

Our approach combines the robustness, high-precision and repeatability of position-controlled industrial robots with compliant control. The requirements and challenges that arise in bimanual manipulation have been covered.

Through the integration of manipulation primitives, workspace manipulability optimization, collision-free motion planning, external wrenches estimation and position-based explicit force control, we achieved a highly dexterous task: bimanual pin insertion.

Future works will include the use of 3D perception systems suitable for industrial applications and the fusion of perception and force information to improve the exploration phase described in \sref{sub:compliant_insertion}.
On the bimanual collision-free motion planning, the use of coordinated motions promises to reduce the time to task completion. Additional work needs to be done in regards of compliant controllers for bimanual collaborative manipulation.
Finally, this work will continue until completion of all the tasks required for assembling an IKEA chair.

\IEEEtriggeratref{10}
\IEEEtriggercmd{\enlargethispage{-115mm}}
\bibliographystyle{IEEEtran}
\bibliography{IEEEabrv,references}

\begin{thebibliography}{10}
\providecommand{\url}[1]{#1}
\csname url@rmstyle\endcsname
\providecommand{\newblock}{\relax}
\providecommand{\bibinfo}[2]{#2}
\providecommand\BIBentrySTDinterwordspacing{\spaceskip=0pt\relax}
\providecommand\BIBentryALTinterwordstretchfactor{4}
\providecommand\BIBentryALTinterwordspacing{\spaceskip=\fontdimen2\font plus
\BIBentryALTinterwordstretchfactor\fontdimen3\font minus
  \fontdimen4\font\relax}
\providecommand\BIBforeignlanguage[2]{{%
\expandafter\ifx\csname l@#1\endcsname\relax
\typeout{** WARNING: IEEEtran.bst: No hyphenation pattern has been}%
\typeout{** loaded for the language `#1'. Using the pattern for}%
\typeout{** the default language instead.}%
\else
\language=\csname l@#1\endcsname
\fi
#2}}

\bibitem{Bischoff2010}
R.~Bischoff, J.~Kurth, G.~Schreiber, R.~Koeppe, A.~Albu-Schaeffer, A.~Beyer,
  O.~Eiberger, S.~Haddadin, A.~Stemmer, G.~Grunwald, and G.~Hirzinger,
  ``\BIBforeignlanguage{English}{{The KUKA-DLR Lightweight Robot arm - a new
  reference platform for robotics research and manufacturing}},'' in
  \emph{\BIBforeignlanguage{English}{6th Ger. Conf. Robot.}}\hskip 1em plus
  0.5em minus 0.4em\relax VDE, 2010, pp. 1--8.

\bibitem{Rooks2006}
B.~Rooks, ``\BIBforeignlanguage{en}{{The harmonious robot}},''
  \emph{\BIBforeignlanguage{en}{Ind. Robot An Int. J.}}, vol.~33, no.~2, pp.
  125--130, Mar. 2006.

\bibitem{Lee1993}
C.~Lee, S.~Chan, and D.~Mital, ``\BIBforeignlanguage{English}{{A joint torque
  disturbance observer for robotic assembly}},'' in
  \emph{\BIBforeignlanguage{English}{Proc. 36th Midwest Symp. Circuits
  Syst.}}\hskip 1em plus 0.5em minus 0.4em\relax IEEE, 1993, pp. 1439--1442.

\bibitem{Knepper2013}
R.~A. Knepper, T.~Layton, J.~Romanishin, and D.~Rus,
  ``\BIBforeignlanguage{English}{{IkeaBot: An autonomous multi-robot
  coordinated furniture assembly system}},'' in
  \emph{\BIBforeignlanguage{English}{IEEE Int. Conf. Robot. Autom.}}\hskip 1em
  plus 0.5em minus 0.4em\relax IEEE, May 2013, pp. 855--862.

\bibitem{Wahrburg2014}
A.~Wahrburg, S.~Zeiss, B.~Matthias, and H.~Ding,
  ``\BIBforeignlanguage{English}{{Contact force estimation for robotic assembly
  using motor torques}},'' in \emph{\BIBforeignlanguage{English}{IEEE Int.
  Conf. Autom. Sci. Eng.}}\hskip 1em plus 0.5em minus 0.4em\relax IEEE, Aug.
  2014, pp. 1252--1257.

\bibitem{Cutkosky1990}
M.~R. Cutkosky and R.~D. Howe, ``{Human grasp choice and robotic grasp
  analysis},'' in \emph{Dextrous Robot Hands}, S.~T. Venkataraman and
  T.~Iberall, Eds.\hskip 1em plus 0.5em minus 0.4em\relax New York, NY:
  Springer New York, 1990, pp. 5--31.

\bibitem{Feix2009}
T.~Feix, R.~Pawlik, H.~Schmiedmayer, J.~Romero, and D.~Kragic, ``{A
  comprehensive grasp taxonomy},'' in \emph{Robot. Sci. Syst. Work. Underst.
  Hum. Hand Adv. Robot. Manip.}, 2009.

\bibitem{Bullock2012}
I.~M. Bullock, R.~R. Ma, and A.~M. Dollar, ``\BIBforeignlanguage{English}{{A
  hand-centric classification of human and robot dexterous manipulation.}}''
  \emph{\BIBforeignlanguage{English}{IEEE Trans. Haptics}}, vol.~6, no.~2, pp.
  129--44, Jan. 2012.

\bibitem{Owen-Hill2013}
A.~Owen-Hill, J.~Bre\~{n}osa, M.~Ferre, J.~Artigas, and R.~Aracil,
  ``\BIBforeignlanguage{en}{{A Taxonomy for Heavy-Duty Telemanipulation Tasks
  Using Elemental Actions}},'' \emph{\BIBforeignlanguage{en}{Int. J. Adv.
  Robot. Syst.}}, vol.~10, no. 371, pp. 1--7, Oct. 2013.

\bibitem{Smith2012}
C.~Smith, Y.~Karayiannidis, L.~Nalpantidis, X.~Gratal, P.~Qi, D.~V.
  Dimarogonas, and D.~Kragic, ``{Dual arm manipulation—A survey},''
  \emph{Rob. Auton. Syst.}, vol.~60, no.~10, pp. 1340--1353, Oct. 2012.

\bibitem{Bloss2010}
R.~Bloss, ``\BIBforeignlanguage{en}{{Robotics innovations at the 2009 Assembly
  Technology Expo}},'' \emph{\BIBforeignlanguage{en}{Ind. Robot An Int. J.}},
  vol.~37, no.~5, pp. 427--430, Aug. 2010.

\bibitem{Yamada1995}
Y.~Yamada, S.~Nagamatsu, and Y.~Sato,
  ``\BIBforeignlanguage{English}{{Development of multi-arm robots for
  automobile assembly}},'' in \emph{\BIBforeignlanguage{English}{IEEE Int.
  Conf. Robot. Autom.}}, vol.~3.\hskip 1em plus 0.5em minus 0.4em\relax IEEE,
  1995, pp. 2224--2229.

\bibitem{Kock2011}
S.~Kock, T.~Vittor, B.~Matthias, H.~Jerregard, M.~Kallman, I.~Lundberg,
  R.~Mellander, and M.~Hedelind, ``\BIBforeignlanguage{English}{{Robot concept
  for scalable, flexible assembly automation: A technology study on a harmless
  dual-armed robot}},'' in \emph{\BIBforeignlanguage{English}{IEEE Int. Symp.
  Assem. Manuf.}}\hskip 1em plus 0.5em minus 0.4em\relax IEEE, May 2011, pp.
  1--5.

\bibitem{Quigley2009}
M.~Quigley, K.~Conley, B.~Gerkey, J.~Faust, T.~Foote, J.~Leibs, E.~Berger,
  R.~Wheeler, and A.~Ng, ``{ROS: an open-source Robot Operating System},'' in
  \emph{ICRA Work. Open Source Softw.}, 2009.

\bibitem{Yoshikawa1985}
T.~Yoshikawa, ``{Manipulability of Robotic Mechanisms},'' \emph{Int. J. Rob.
  Res.}, vol.~4, no.~2, pp. 3--9, June 1985.

\bibitem{Dubey1995}
R.~Dubey, ``\BIBforeignlanguage{English}{{A weighted least-norm solution based
  scheme for avoiding joint limits for redundant joint manipulators}},''
  \emph{\BIBforeignlanguage{English}{IEEE Trans. Robot. Autom.}}, vol.~11,
  no.~2, pp. 286--292, Apr. 1995.

\bibitem{Kuffner2000}
J.~Kuffner and S.~LaValle, ``\BIBforeignlanguage{English}{{RRT-connect: An
  efficient approach to single-query path planning}},'' in
  \emph{\BIBforeignlanguage{English}{IEEE Int. Conf. Robot. Autom.}},
  vol.~2.\hskip 1em plus 0.5em minus 0.4em\relax IEEE, 2000, pp. 995--1001.

\bibitem{diankov2010}
R.~Diankov, ``{Automated Construction of Robotic Manipulation Programs},''
  Ph.D. dissertation, Carnegie Mellon University, Robotics Institute, Aug.
  2010.

\bibitem{Lavalle2006}
S.~M. LaValle, \emph{{Planning algorithms}}.\hskip 1em plus 0.5em minus
  0.4em\relax Cambridge university press, 2006.

\bibitem{Swevers2007}
J.~Swevers, W.~Verdonck, and J.~{De Schutter},
  ``\BIBforeignlanguage{English}{{Dynamic Model Identification for Industrial
  Robots}},'' \emph{\BIBforeignlanguage{English}{IEEE Control Syst. Mag.}},
  vol.~27, no.~5, pp. 58--71, Oct. 2007.

\bibitem{Kubus2008}
D.~Kubus, T.~Kroger, and F.~Wahl, ``\BIBforeignlanguage{English}{{On-line
  estimation of inertial parameters using a recursive total least-squares
  approach}},'' in \emph{\BIBforeignlanguage{English}{IEEE/RSJ Int. Conf.
  Intell. Robot. Syst.}}\hskip 1em plus 0.5em minus 0.4em\relax IEEE, Sept.
  2008, pp. 3845--3852.

\bibitem{Hollerbach2008}
J.~Hollerbach, W.~Khalil, and M.~Gautier, ``{Model Identification},'' in
  \emph{Springer Handb. Robot.}, B.~Siciliano and O.~Khatib, Eds.\hskip 1em
  plus 0.5em minus 0.4em\relax Springer, 2008, ch.~1, pp. 321--344.

\bibitem{Ott2010}
C.~Ott, R.~Mukherjee, and Y.~Nakamura, ``\BIBforeignlanguage{English}{{Unified
  Impedance and Admittance Control}},'' in
  \emph{\BIBforeignlanguage{English}{IEEE Int. Conf. Robot. Autom.}}\hskip 1em
  plus 0.5em minus 0.4em\relax IEEE, May 2010, pp. 554--561.

\bibitem{Seraji1994}
H.~Seraji, ``\BIBforeignlanguage{English}{{Adaptive admittance control: an
  approach to explicit force control in compliant motion}},'' in
  \emph{\BIBforeignlanguage{English}{IEEE Int. Conf. Robot. Autom.}}\hskip 1em
  plus 0.5em minus 0.4em\relax IEEE Comput. Soc. Press, 1994, pp. 2705--2712.

\bibitem{Calanca2015}
A.~Calanca, R.~Muradore, and P.~Fiorini, ``{A Review of Algorithms for
  Compliant Control of Stiff and Fixed-Compliance Robots},'' \emph{IEEE/ASME
  Trans. Mechatronics}, vol.~PP, no.~99, p.~1, 2015.

\bibitem{Hannaford1991}
B.~Hannaford, L.~Wood, D.~A. McAffee, and H.~Zak, ``{Performance evaluation of
  a six-axis generalized force-reflecting teleoperator},'' \emph{IEEE Trans.
  Syst. Man Cybern.}, vol.~21, no.~3, pp. 620--633, 1991.

\bibitem{Yip2011}
M.~C. Yip, M.~Tavakoli, and R.~D. Howe, ``{Performance Analysis of a Haptic
  Telemanipulation Task under Time Delay},'' \emph{Adv. Robot.}, vol.~25,
  no.~5, pp. 651--673, Jan. 2011.

\end{thebibliography}

\end{document}